\documentclass{llncs}
\usepackage{epsf}

\usepackage{times}
\usepackage[hyphens]{url}
\usepackage{paralist}
\usepackage{amsmath}
\usepackage{pgfplots}
\usepackage{color}

\usepackage{amssymb}
\usepackage{algorithm}
\usepackage{algpseudocode}
\usepackage{xspace}
\usepackage{framed}

\usetikzlibrary{patterns}

\defaultleftmargin{0em}{}{}{}

\begin{document}

\title{A Parallel Corpus of Translationese}

\author{Ella Rabinovich  \and Shuly Wintner \and Ofek Luis Lewinsohn}
\institute{
  E.~Rabinovich, S.~Wintner \\ Department of Computer Science, University of Haifa. \\{ellarabi@csweb.haifa.ac.il,\quad shuly@cs.haifa.ac.il} \and
  O.L.~Lewinsohn \\ Department of Computational Linguistics, Universit{\"a}t des Saarlandes. \\{o.l.lewinsohn@gmail.com}
}

\maketitle

\begin{abstract}
We describe a set of bilingual English--French and English--German parallel corpora in which the direction of translation is accurately and reliably annotated. The corpora are diverse, consisting of parliamentary proceedings, literary works, transcriptions of TED talks and political commentary. They will be instrumental for research of translationese and its applications to (human and machine) translation; specifically, they can be used for the task of translationese identification, a research direction that enjoys a growing interest in recent years. To validate the quality and reliability of the corpora, we replicated previous results of supervised and unsupervised identification of translationese, and further extended the experiments to additional datasets and languages.
\end{abstract}

\section{Introduction}
\label{sec:introduction}

Research in all areas of language and linguistics is stimulated by the unprecedented availability of data.
In particular, large text corpora are essential for research of the unique properties of \emph{translationese}: the sub-language of translated texts (in any given language) that is presumably distinctly different from the language of texts originally written in the same language. Indeed, contemporary research in translation studies is prominently dominated by corpus-based approach \cite{Baker:1993,Baker:1995,Baker:1996,Al-Shabab:1996,Laviosa:1998,laviosa2002corpus,olohan:2004,becher:2011,Zanettin201320}. Most studies of translationese utilize \emph{monolingual comparable} corpora, i.e. corpora where translations from multiple languages into a single language are compared with texts written originally in the target language.

The unique characteristics of translated language have been traditionally classified into two categories: properties that stem from the \emph{interference} of the source language \cite{Gellerstam:1986}, and \emph{universal} traits elicited from the translation process itself, regardless of the specific source and target language-pair \cite{Baker:1993,Toury:1995}. Computational investigation of translated texts has been a prolific field of recent research, laying out an empirical foundation for the theoretically-motivated hypotheses on the characteristics of translationese. More specifically, identification of translated texts by means of automatic classification shed much light on the manifestation of translation universals and interference phenomena in translation \cite{Baroni2006,Halteren08,Kurokawa:etal:2009,koppel-ordan:2011:ACL-HLT2011,Ilisei2010,Ilisei:2012,vered:noam:shuly,TACL618,nisioi2015b}.

Along the way it was suggested that the unique properties of translationese should be studied in a \emph{parallel} setting, i.e., in context of the corresponding \emph{source} language: the original language the text was produced in. In particular, several studies hypothesized that certain phenomena traditionally attributed to translation universals (i.e., source-language independent) are, in fact, derivatives of the linguistic characteristics of the specific language-pair, subject to translation. \cite{Pym:2008} investigated the phenomenon of omitting the optional ``that'' in reporting English verbs, such as ``he claimed [that] they left the room'', highlighting correlation of this behavior to the linguistic conventions in the source language. \cite{becher:2010} raised similar arguments regarding \emph{explicitation} in translation (e.g., excessive usage of cohesive devices): he claimed that this phenomenon should only be studied by comparative analysis of translation and its original counterpart.

Parallel setting can also facilitate the task of automatic identification of translationese. \cite{Kurokawa:etal:2009} were the first to employ bilingual text properties for identification  of the direction of translated parallel texts. They took advantage of sentence pairs translated in both directions for training a supervised classifier to identify translationese using word- and (part-of-speech) POS-ngrams as features. \cite{eetemadi-toutanova:2014} used POS MTU (minimal translation unit) ngrams and HMM distortion properties extracted from bilingual parallel English-French Europarl and Hansard texts. Consecutively, they carried out series of experiments on sentence-level identification of translationese using Brown clusters \cite{Brown92classbased} MTUs on the Hansard corpus \cite{eetemadi:toutanova:2015}.

A good corpus for research into the properties of translationese, should ideally satisfy the following desiderata:
\begin{compactdesc}
\item[Diversity] The corpus should ideally reflect diverse genres, registers, authors, modality (written vs. spoken) etc.
\item[Parallelism] The corpus should include both the source and its translation, so that features that are revealed in the translation can be traced back to their origins in the source.
\item[Multilinguality] Having translations from several source languages to the same target language facilitates a closer inspection of properties that are language-pair-specific vs.\ more ``universal'' features of translationese \cite{Baker:1993,house:2008,laviosa:2009,becher:2010}.
\item[Uniformity] Whatever processing is done on the texts, it must be done uniformly. This includes sentence boundary detection, tokenization, sentence- and word-alignment, POS tagging, etc.
\item[Availability] Finally, corpora that are used for research must be publicly available so that other researchers have the opportunity to replicate and corroborate research results.
\end{compactdesc}

In this work we describe a set of cross-domain, parallel, uniform, English-French and English-German corpora that were compiled specifically for research on translationese\footnote{All corpora are available at \url{http://cl.haifa.ac.il/projects/translationese/index.shtml}}. The corpora are diverse, consisting of parliamentary proceedings, literary works, transcriptions of TED talks and political commentary. We rigorously evaluate all datasets by series of supervised and unsupervised experiments; sensitivity analysis further implies applicability of these methodologies to data-meager scenarios. These datasets will be instrumental for research of translationese and its manifestations; they will also facilitate accurate identification of translationese at small text units by exploitation of bilingual text properties.

We detail the structure of the corpus in Section~\ref{sec:structure}, explain how it was processed in Section~\ref{sec:processing}, and evaluate it by extending some state-of-the-art supervised and unsupervised experiments in Section~\ref{sec:evaluation}. We conclude with suggestions for future extensions.

\section{Corpus structure}
\label{sec:structure}
Our corpus of translationese consists of five sub-corpora: Europarl, Canadian Hansard, literature, TED and political commentary.\footnote{We use ``EUR'', ``HAN'', ``LIT'', ``TED'' and ``POL'' to denote the five corpora hereafter.} All are parallel corpora, with accurate annotation indicating the direction of the translation. The datasets are uniformly pre-processed, represented, and organized.  All corpora were further filtered to contain solely one-to-one sentence-alignments, which are more useful for the SMT research. Tables~\ref{tbl:corpus-stats-en-fr} and~\ref{tbl:corpus-stats-en-de} report some statistical data on the corpus (after tokenization).

\begin{table*}[hbt]
\centering
\caption{English-French corpus statistics}
\label{tbl:corpus-stats-en-fr}
\begin{tabular}{l|rrr|rr|rr}
\hline
& \multicolumn{3}{c|}{\# of sentences} & \multicolumn{2}{c|}{\# of tokens} & \multicolumn{2}{c}{\# of types} \\
\hline
\multicolumn{1}{c|}{corpus} & \multicolumn{1}{c}{Original EN} & \multicolumn{1}{c}{Original FR} & \multicolumn{1}{c|}{total} & \multicolumn{1}{c}{Original EN} & \multicolumn{1}{c|}{Original FR} & \multicolumn{1}{c}{Original EN} & \multicolumn{1}{c}{Original FR} \\
\hline
EUR & 217K & 130K & 347K & 9,572K & 10,542K & 61K & 73K \\
HAN & 5,237K & 1,379K & 6,616K & 132,232K & 147,463K & 193K & 196K \\
LIT & 35K & 98K & 133K & 2,875K & 2,898K & 52K & 66K \\
TED & 7K & 4K & 12K & 217K & 239K & 14K & 17K \\
\hline
\end{tabular}
\end{table*}

\begin{table*}[hbt]
\centering
\caption{English-German corpus statistics}
\label{tbl:corpus-stats-en-de}
\begin{tabular}{l|rrr|rr|rr}
\hline
& \multicolumn{3}{c|}{\# of sentences} & \multicolumn{2}{c|}{\# of tokens} & \multicolumn{2}{c}{\# of types} \\
\hline
\multicolumn{1}{c|}{corpus} & \multicolumn{1}{c}{Original EN} & \multicolumn{1}{c}{Original DE} & \multicolumn{1}{c|}{total} & \multicolumn{1}{c}{Original EN} & \multicolumn{1}{c|}{Original DE} & \multicolumn{1}{c}{Original EN} & \multicolumn{1}{c}{Original DE} \\
\hline
EUR & 225K & 155K & 380K & 10,550K & 10,067K & 63K & 170K \\
LIT & 45K & 48K & 93K & 2,854K & 2,666K & 56K & 104K \\
POL & 8K & 9K & 18K & 443K & 421K & 26K & 44K \\
\hline
\end{tabular}
\end{table*}

\subsection{Europarl}
\label{sec:europarl}
The Europarl  sub-corpus is extracted from the collection of the proceedings of the European Parliament, dating back to 1996, originally collected by \cite{Koehn05Europarl}. The original corpus\footnote{The original Europarl is available from \url{http://www.statmt.org/europarl/}} is organized as several language-pairs, each with multiple sentence-aligned files. We mainly used the English-French and English-German segments, but resorted to other segments as we presently explain. We focus below on the way we generated the English-French sub-corpus; its English-German counterpart was obtained in a similar way.

Europarl is probably the most popular parallel corpus in natural language processing, and it was indeed used for many of the translationese tasks surveyed in Section~\ref{sec:introduction}. Unfortunately, it is a very problematic corpus. First, it consists of transcriptions of spoken utterances that are edited (by the speakers) after they are transcribed; only then are they translated. Consequently, there are significant discrepancies between the actual speeches and their ``verbatim'' transcriptions \cite{cucchi:2012}. Second, while ``Members of the European Parliament have the right to use any of the EU's [24] official languages when speaking in Parliament'',\footnote{\url{http://europa.eu/about-eu/facts-figures/administration/index_en.htm}} many of them prefer to speak in English, which is often not their native language.\footnote{\url{http://www.theguardian.com/education/datablog/2014/may/21/european-parliament-english-language-official-debates-data}}

Mainly due to its multilingual nature, however, Europarl has been used extensively in SMT \cite{koehn-birch-steinberger:2010:MT-Summit-XII} and in cross-lingual research \cite{Cartoni_BJL_2013}. It has even been adapted specifically for research in translation studies: \cite{ISLAM12.729} compiled a customized version of Europarl, where the direction of translation is indicated. They used meta-data from the corpus, and in particular the ``language'' tag, to identify the original language in which each sentence was spoken, and removed sentence pairs for which this information was missing. A similar strategy was used by \cite{lembersky-ordan-wintner:CL2012} and \cite{CARTONI12.188}. However, relying on the ``language'' tag in Europarl parallel text for identification of translation direction could be potentially flawed. Next we detail the procedure of reliable extraction of speaker details, including the original language of each sentence.

The Europarl corpus is a collection of several monolingual (parallel) corpora: the original text was uttered in one language and then translated to several other languages. In each sub-corpus, each paragraph is annotated with meta-information, in particular, the original language in which the paragraph was uttered. Unfortunately, the meta-information pertaining to the original language of Europarl utterances is frequently missing. Furthermore, in some cases this information is inconsistent: different languages are indicated as the original languages of (various translations of) the same paragraph (in the various sub-corpora). Additionally, the Europarl corpus includes several bilingual sub-corpora that are generated from the original and the translated texts, and are already sentence-aligned. These bilingual corpora include only raw sentence pairs, with no meta-information.

To minimize the risk of erroneous information, we processed the Europarl corpus as follows. First, we propagated the meta-information from the monolingual texts to the bilingual sub-corpora: each sentence pair was thus annotated with the original language in which it was uttered. We repeated this process five times, using as the source of meta-information the original monolingual corpora in five languages: English, French, German, Italian, and Spanish (note that not all monolingual corpora are identical: some are much larger than others). For the same reason, not all the English-French sentence pairs in our bilingual corpus are reflected in all five monolingual corpora, and therefore some sentence pairs have less than five annotations of the original language. We restricted the bilingual corpus to only those sentence pairs that had five annotations. Then, we filtered out all sentence pairs whose annotations were inconsistent (about 0.5\%). We also removed comments (about 0.5\% as well), typically written in parentheses (things like ``applause'', ``continuation of the previous session'', etc.) As a result, we are confident that the speaker information (and in particular, the original language of utterances) in the filtered corpus is highly accurate.

\subsection{The Canadian Hansard}
The Hansard corpus is a parallel corpus consisting of transcriptions of the Canadian parliament in (Canadian) English and French from 2001--2009. We used a version that was annotated with the original language of each parallel sentence.  This corpus most likely suffers from similar problems as the Europarl corpus discussed above; indeed, \cite{doi:10.3366/cor.2007.2.2.187}, who investigated the \emph{British} Hansard parliamentary transcripts, found that ``the transcripts omit performance characteristics of spoken language, such as incomplete utterances or hesitations, as well as any type of extrafactual, contextual talk'' and that ``transcribers and editors also alter speakers' lexical and grammatical choices towards more conservative and formal variants.'' Still, this is the largest available source of English--French sentence pairs. In addition to parliament members' speech, the original Hansard corpus contains metadata. Various annotations were used to discriminate different line types, including the date of the session, the name of the speaker, etc. We filtered out all segments except those referring to speech: in total, about 15\% of the corpus line-pairs were thus eliminated.

\subsection{Literary Classics}
Our English--French literary corpus consists of classics written and translated in the 18th--20th centuries by English and French writers. Most of the raw material is available from the Gutenberg project\footnote{\url{http://www.gutenberg.org}} and FarkasTranslations.\footnote{\url{http://farkastranslations.com/}} 
The English-German literature corpus was generated in a similar way: we used material from the Gutenberg project, Wikisource,\footnote{\url{http://en.wikisource.org/}} and a few more books. Both English--French and English--German datasets contain a metadata file with details about the books: title, year of publication, translator name  and year of translation.

Identification of translationese in literary text by means of classification is considered a more challenging task \cite{LynchV12,Udi2013} than classifying more ``technical'' translations, such as parliament proceedings. Translators of literature typically benefit from freedom and fewer constraints, rendering the translated text more similar to original writing. Additionally, our literature span almost three centuries and comprises works from wide range of genres -- traits that overshadow the subtle characteristics of translationese \cite{koppel-ordan:2011:ACL-HLT2011,Popescu11,udi:llc,TACL618}. Under this circumstances, we obtain very high accuracy with supervised classification on this corpus, and moderate, yet reasonable results with unsupervised clustering (see Section~\ref{sec:results}).

\subsection{TED Talks}
Our TED corpus is based on the subtitles of the TED talks delivered in English and translations to English of TEDx talks originally given in French\footnote{TEDx are TED-like events not restricted to specific language. We could not find sufficient amount of TEDx German talks translated to English.}. We used the TED API\footnote{\url{http://developer.ted.com/}} to extract subtitles of talks delivered in English, and Youtube API for TEDx talks originally given in French.

The quality of translation in this corpus is very high: not only are translators assumed to be competent, but the common practice is that each translation passes through a review before being published. This corpus consists of talks delivered orally, but we assume that they were meticulously prepared, so the language is not spontaneous but rather planned. Compared to the other sub-corpora, the TED dataset has some unique characteristics that stem from the following reasons:
\begin{inparaenum}[(i)]
\item its size is relatively small;
\item it exhibits stylistic disparity between the original and translated texts (the former contains more ``oral'' markers of a spoken language, while the latter is a written translation); and
\item TED talks are not transcribed but are rather subtitled, so they undergo some editing and rephrasing. 
\end{inparaenum}
The vast majority of TED talks are publicly available online, which makes this corpus easily extendable for future research.

\subsection{Political News and Commentary}
This corpus contains articles, commentary and analysis on world affairs and international relations. English articles and their translations to German were collected from Project Syndicate.\footnote{\url{http://www.project-syndicate.org/}} This is a non-profit organization that primarily relies on contributions from newspapers in developed countries. It provides original commentaries by people who are shaping the world's economics, politics, science and culture. We collected articles categorized as Word Affairs from this project, originally written by English authors and translated to German. Original German commentaries and their translations to English were collected from the Diplomatics Magazine,\footnote{\url{http://www.diplomatisches-magazin.de/}} specifically, from its International Relations section.

\section{Processing}
\label{sec:processing}
The original Europarl corpora are already sentence-aligned, using an implementation of the Gale and Church sentence-alignment algorithm \cite{972455}. Since the alignment was done for one source paragraph at a time (typically consisting of few sentences), its quality is very high. The same also holds for the Hansard corpus, so we used the original alignments for both sub-corpora. We then filtered out any alignments that were not one-to-one; this resulted in a loss of about 3\% of the alignments in Europarl, and only 2\% in Hansard.

The literary sub-corpus required more careful attention. Books that were acquired from \url{FarkasTranslations.com} were available pre-aligned at the chapter- and paragraph-level; we therefore sentence-aligned them, one paragraph at a time, using a Python implementation \cite{tan:bond:2014} of the Gale and Church algorithm. For the remainder of the books, we first extracted chapters by (manually) identifying characteristic chapter titles (e.g., Roman numerals, explicit ``Chapter N''). Paragraph boundaries within a chapter are typically marked by a double newline in Gutenberg transcripts, and we used this pattern to break chapters into paragraphs. Due to the fact that the Gale-Church algorithm only utilizes text length for alignment, it can be easily refined for aligning other logical units, e.g., paragraphs \cite{972455}. Finally, we aligned sentences within paragraphs using the same algorithm.

The genre of the literature sub-corpus is very different (presumably due to translators taking greater liberty), hence restricting the dataset to include only one-to-one sentence-alignments resulted in loss of above 10\% of each book.

Sentence-alignment of subtitles of TED talks originally delivered in French (and translated to English) involved synchronization of subtitle frames. A typical frame in a subtitles (.srt) file contains frame start and end time (including milliseconds), as well as frame text:

\begin{center}
\begin{framed}
\begin{tabular}{@{\hskip 0em}l@{\hskip -5em}r}
\verb+18+ &  frame sequential number \\
\verb+00:00:47,497 --> 00:00:50,813+ &  frame start and end time \\
\verb+Cet engagement, je pense que j'ai fait le choix+ & frame text
\end{tabular}
\end{framed}
\end{center}

First, we re-organized the subtitles file to contain (longer) frames that start and end on a sentence boundary; we achieved this by concatenating frames until a sentence termination punctuation symbol is reached. This procedure was conducted on both French subtitles and their corresponding English translations. Then, we aligned the English--French parallel files at paragraph-level by alternated concatenation of paragraphs until synchronization of frame end time (up to a $\delta$ threshold that was fixed to 500 milliseconds) on the English and French sides. The paragraph-alignment procedure pseudo-code is detailed in Algorithm~\ref{fig:ted-alignment}.

\begin{algorithm}[!htb]
\caption{TED subtitles paragraph-alignment algorithm}
\label{fig:ted-alignment}
\begin{algorithmic}
\State Comment: $l\_paragraphs$ and $r\_paragraphs$ are (not necessarily equal length)
arrays of text paragraphs for alignment, from the left and right sides, respectively
\smallskip
\State $\delta$ = 500 milliseconds \Comment{threshold controlling the allowed delta in aligned frames' end time}
\smallskip
\State subtitles\_paragraph\_alignment(1,1) \Comment{initial invocation assuming the arrays start from 1}
\smallskip
\Procedure{\textnormal{subtitles\_paragraph\_alignment($l\_count$,$r\_count$)}} {}
\If {($l\_count$ \textgreater \xspace $l\_paragraphs.length$) \textbf{and} ($r\_count$ \textgreater \xspace $r\_paragraphs.length$)}
\State return
\EndIf
\If {($l\_count$ \textgreater \xspace $l\_paragraphs.length$)}
\State output $r\_paragraphs$[$r\_count$:$r\_paragraphs.length$] \Comment remainder of the right side
\State return
\EndIf
\smallskip
\If {($r\_count$ \textgreater \xspace $r\_paragraphs.length$)}
\State output $l\_paragraphs$[$l\_count$:$l\_paragraphs.length$] \Comment remainder of the left side
\State return
\EndIf
\smallskip
\State $l\_current$ = $l\_paragraphs$[$l\_count$].$frame\_content$
\State $l\_frame\_end$ = $l\_paragraphs$[$l\_count$].$frame\_end$
\State $r\_current$ = $r\_paragraphs$[$r\_count$].$frame\_content$
\State $r\_frame\_end$ = $r\_paragraphs$[$r\_count$].$frame\_end$
\smallskip
\While {($\lvert l\_frame\_end$\xspace\xspace --\xspace$r\_frame\_end \lvert$ \textgreater\xspace $\delta$ \textbf{and}
($l\_count$ \textless \xspace $l\_paragraphs.length$)
\State \textbf{and} ($r\_count$ \textless \xspace $r\_paragraphs.length$))}
\If {($l\_frame\_end$ \textgreater\xspace $r\_frame\_end$)} \Comment advance on the right side
\State $r\_count$ += 1; $r\_current$ += $r\_paragraphs$[$r\_count$].$frame\_content$
\State $r\_frame\_end$ = $r\_paragraphs$[$r\_count$].$frame\_end$
\Else \Comment advance on the left side
\State $l\_count$ += 1; $l\_current$ += $l\_paragraphs$[$l\_count$].$frame\_content$
\State $l\_frame\_end$ = $l\_paragraphs$[$l\_count$].$frame\_end$
\EndIf
\EndWhile
\smallskip
\State output ``aligned paragraph pair:'', $l\_current$, $r\_current$
\State subtitles\_paragraph\_alignment($l\_count$+1,$r\_count$+1) \Comment recursive invocation
\EndProcedure
\end{algorithmic}
\end{algorithm}

We further aligned the paragraph-aligned TED and TEDx corpora at the sentence-level using the same sentence-alignment procedure \cite{972455}. TED talks tend to vary greatly in terms of sentence alignments (one-to-one, one-to-many, many-to-one, many-to-many). On average, approximately~10\% of the alignments are not one-to-one; those were filtered out as well.


\section{Evaluation}
\label{sec:evaluation}
To validate the quality of the corpus we replicated the experiments of \cite{vered:noam:shuly}, who conducted a thorough exploration of supervised classification of translationese, using dozens of feature types. While \cite{vered:noam:shuly} only used the Europarl corpus (in its original format) and worked on English translated from French, we extended the experiments to all the datasets described above, including also English translated from German, as well as French and German translations from English. We show that in-domain classification (with ten-fold cross-validation evaluation) yields excellent results. Moreover, very good results are obtained using unsupervised classification, implying robustness of this methodology and its applicability to various domains and languages.

\subsection{Preprocessing and tools}
\label{sec:tools}
The (tokenized) datasets were split into chunks of approximately 2000 tokens, respecting sentence boundaries and preserving punctuation. We assume that translationese features are present in the texts or speeches across author, genre or native language, thus we allow some chunks to contain linguistic information from two or more different speakers simultaneously. The frequency-based features are normalized by the number of tokens in each chunk. For POS tagging, we employ the Stanford implementation along with its models for English, French and German \cite{manning-EtAl:2014:P14-5}.

We use Platt's sequential minimal optimization algorithm \cite{Keerthi2001} to train a support vector machine classifier with the default linear kernel, an implementation freely available in Weka \cite{weka}. In all classification experiments we use (the maximal) equal number of chunks from each class: original (O) and translated (T).

\subsection{Features}
\label{sec:features}
The first feature set we utilized for classification tasks comprises \emph{function words} (FW), probably the most popular choice ever since \cite{mostWallace} used it successfully for the Federalist Papers. Function words proved to be suitable features for multiple reasons:%
\begin{inparaenum}[(i)]
\item they abstract away from contents and are therefore less biased by topic;
\item their frequency is so high that by and large they are assumed to be selected unconsciously by authors;
\item although not easily interpretable, they are assumed to reflect grammar, and therefore facilitate the study of how structures are carried over from one language to another.
\end{inparaenum}
We used the list of above 400 English function words provided in \cite{koppel-ordan:2011:ACL-HLT2011}, and similar number of French and German function words.\footnote{The list of French and German FW was downloaded from \url{https://code.google.com/archive/p/stop-words/}.}

A more informative way to represent (admittedly shallow) syntax is to use \emph{part-of-speech (POS) trigrams}.
Triplets such as PP (personal pronoun) + VHZ (verb ``have'', 3rd person sing. present) + VBN (verb ``be'', past participle) reflect a complex tense form, represented distinctively across languages. In Europarl, for example, this triplet is highly frequent in translations from Finnish and Danish and much rarer in translations from Portuguese and Greek.

We also used \emph{positional token frequency} \cite{Grieve:2007}. The feature is defined as counts of words occupying the first, second, third, penultimate and last positions in a sentence. The motivation behind this feature is that sentences open and close differently across languages, and it should be expected that these opening and closing devices will be transferred from the source if they do not violate the grammaticality of the target language. Positional tokens were previously used for translationese identification \cite{vered:noam:shuly} and for native language detection \cite{nisioi2015a}.

Finally, we experimented with \emph{contextual function words}. Contextual FW are a variation of POS trigrams where a trigram can be anchored by specific function words: these are consecutive triplets $\langle w_1$,$w_2$,$w_3 \rangle$ where at least two of the elements are function words, and at most one is a POS tag.

POS-trigrams, positional tokens and contextual-FW-trigrams are calculated as detailed in \cite{vered:noam:shuly}, but we only considered the 1000 most frequent feature values extracted from each dataset. 

\subsection{Results}
\label{sec:results}
\subsubsection{Supervised identification of translationese}

We begin with supervised identification of translated text using features detailed in Section~\ref{sec:features}\footnote{Feature combinations yield similar, occasionally slightly better, results; we refrain from providing full analysis in this paper.}; table~\ref{tbl:supervised-cl} reports the results. Total number of chunks used for classification is reported per dataset, where we used the maximum available amount of data (up to 1000 chunks).

\begin{table}[hbt]
\caption{Ten-fold supervised cross-validation classification (rounded) accuracy of English, French and German translationese; the best result in each column is boldfaced.}
\label{tbl:supervised-cl}
\centering
\begin{tabular}{@{}l|cccc|ccc|cccc|ccc}
\hline
& \multicolumn{4}{c|}{EN(O)+FR$\rightarrow$EN} & \multicolumn{3}{c|}{EN(O)+DE$\rightarrow$EN} & \multicolumn{4}{c|}{FR(O)+EN$\rightarrow$FR} & \multicolumn{3}{c}{DE(O)+EN$\rightarrow$DE} \\
\hline
feature / corpus & EUR & HAN & LIT & TED & EUR & LIT & POL & EUR & HAN & LIT & TED & EUR & LIT & POL \\
\hline
total \# of chunks & 1K & 1K & 400 & 40 & 1K & 650 & 100 & 1K & 1K & 400 & 40 & 1K & 600 & 90 \\
\hline
FW              & 96 & \textbf{99} & \textbf{99} & 90 & 96 & \textbf{95} & \textbf{100} & \textbf{96} & 94 & 93 & 93 & 99 & \textbf{96} & 99  \\
pos. tokens     & 97 & 96 & 96 & 93 & 93 & 93 & 99  & \textbf{96} & \textbf{98} & \textbf{95} & 96 & \textbf{98} & 92 & \textbf{100} \\
POS-trigrams    & \textbf{98} & \textbf{99} & 98 & \textbf{94} & \textbf{97} & 94 & \textbf{100} & 95 & 88 & \textbf{95} & 94 & \textbf{98} & 93 & 99  \\
contextual FW   & 94 & 98 & 93 & 90 & 92 & 89 & 98  & \textbf{96} & 95 & 94 & \textbf{98} & 94 & 83 & 90  \\
\hline
\end{tabular}
\end{table}

In line with previous works, the classification results are very high, yielding near-perfect accuracy with all feature types across all datasets. Close inspection of highly discriminative feature values sheds interesting light on the realization of unique characteristics of translationese across languages and domains; we leave this discussion for another venue.

Supervised classification methods, however, suffer from two main drawbacks:\\
\begin{inparaenum}[(i)]
\item they inherently depend on data annotated with the translation direction, and
\item they may not be generalized to unseen (related or unrelated) domains.
\end{inparaenum}
Indeed, series of works on supervised identification of translationese reveal that classification accuracy dramatically deteriorates when classifier is evaluated out-of-domain (i.e., trained and tested on texts drawn from different corpora): \cite{koppel-ordan:2011:ACL-HLT2011,Popescu11,udi:llc,TACL618} demonstrated significant drop in the accuracy of classification when one of the parameters (genre, source language, modality) was changed. These shortcomings undermine the usability of supervised methods for translationese identification in a typical real-life scenario, where no labelled in-domain data are available.

\subsubsection{Unsupervised identification of translationese}
To overcome the domain- and labeled-data-dependence of supervised classification we experiment in this section with unsupervised methods. We adopt the approach detailed in \cite{TACL618}, who demonstrated high accuracy identifying English translationese by clustering methodology.

Table~\ref{tbl:unsupervised-cl} demonstrates the results; the reported numbers reflect average accuracy over 30 experiments (the only difference being a random choice of the initial conditions).\footnote{Standard deviation in most experiments was close to~0.} Europarl and Hansard systematically obtain very high accuracy with all feature types (with a single exception of FW for French Hansard), implying uniform distribution of other linguistic aspects (authorship, topic, modality, epoch etc.) in these sub-corpora, thus facilitating the unsupervised procedure of clustering, since the text translation status dominates other dimensions.

\begin{table}[hbt]
\caption{Clustering (rounded) accuracy of English, French and German translationese; the best result in each column is boldfaced.}
\label{tbl:unsupervised-cl}
\centering
\begin{tabular}{@{}l|cccc|ccc|cccc|ccc}
\hline
& \multicolumn{4}{c|}{EN(O)+FR$\rightarrow$EN} & \multicolumn{3}{c|}{EN(O)+DE$\rightarrow$EN} & \multicolumn{4}{c|}{FR(O)+EN$\rightarrow$FR} & \multicolumn{3}{c}{DE(O)+EN$\rightarrow$DE} \\
\hline
feature / corpus & EUR & HAN & LIT & TED & EUR & LIT & POL
& EUR & HAN & LIT & TED & EUR & LIT & POL \\
\hline
total \# of chunks & 1K & 1K & 400 & 40 & 1K & 650 & 100 & 1K & 1K & 400 & 40 & 1K & 600 & 90 \\
\hline
FW              & 92 & 91 & 77 & \textbf{89} & \textbf{95} & \textbf{70} & \textbf{100} & 91 & 71 & 72 & 95 & \textbf{96} & \textbf{68} & 98  \\
pos. tokens     & 87 & 95 & 55 & 67 & 80 & 64 & 99  & \textbf{97} & 86 & \textbf{80} & 83 & 94 & \textbf{68} & \textbf{99}  \\
POS-trigrams    & \textbf{97} & 94 & 71 & 61 & 94 & 67 & \textbf{100} & 95 & 79 & 60 & 85 & 95 & \textbf{68} & 99  \\
contextual FW   & 87 & \textbf{96} & \textbf{78} & 70 & 88 & 67 & 98  & 96 & \textbf{91} & 72 & \textbf{98} & 95 & 64 & 89  \\
\hline
\end{tabular}
\end{table}

A notably high accuracy is obtained on the small TED corpus, which implies the applicability of the clustering methodology to data-meager scenarios. The exceptionally high accuracy achieved by unsupervised procedure on the politics dataset (both English and German, across all feature types) may indicate existence of additional artifacts (e.g., subtle topical differences) that tease apart O from T, thus boosting the classification procedure. We leave this investigation for the future work.

We explain the lower precision achieved on the literature corpus by its unique character: translators of literary works enjoy more freedom, rendering the translated texts more similar to original writing. Yet, clustering with FW systematically yields a reasonable accuracy for the literature datasets as well. We therefore, conclude that FW comprise one of the best-performing and most-reliable features for the task of unsupervised identification of translationese.

\subsubsection{Sensitivity analysis}
\begin{figure*}[hbt]
\centering
\pgfplotsset{every tick label/.append style={font=\tiny}}
\pgfplotsset{every axis plot/.append style={smooth,black!100,densely dotted}}
\begin{tikzpicture}
\begin{axis}[
    axis lines=middle,
    height=6cm, width=12.5cm,
    xlabel={total number of chunks},
    ylabel={classification accuracy (\%)},
    ylabel near ticks,
    xlabel near ticks,
    xmin=199, xmax=2000,
    ymin=84,  ymax=101,
    xtick={200,400,600,800,1000,1200,1600,2000},
    ytick={85,90,100},
    ymajorgrids=true,
    grid style=dashed,
    legend pos=south east,
    legend style={/tikz/column 2/.style={column sep=5pt,},font=\tiny},
    legend columns=2,
    mark options={solid,black},
    ]
\addplot[mark=star] coordinates             {(200,99)(400,98.5)(600,98.5)(800,98.75)(1000,99)(1200,99.33)(1600,99.37)(2000,99.25)};
\addplot[mark=square] coordinates           {(200,95)(400,96.5)(600,95.1)(800,95.25)(1000,94.19)(1200,94.5)(1600,95.56)(2000,95.55)};
\addplot[mark=o] coordinates                {(200,91.5)(400,93.25)(600,93.67)(800,94.37)(1000,93.9)(1200,95.25)(1600,94.75)(2000,95.7)};
\addplot[mark=+] coordinates                {(200,91.5)(400,92.25)(600,93)(800,95.25)(1000,94.39)(1200,95.91)(1600,95.62)(2000,95.7)};
\addplot[mark=x] coordinates                {(200,98.5)(400,99.25)(600,98.5)(800,98.5)(1000,99.3)(1200,99)(1600,99.4)(2000,98.8)};
\addplot[mark=triangle] coordinates         {(200,86.5)(400,93)(600,92.5)(800,92.75)(1000,94.6)(1200,93.9)(1600,94.25)(2000,94.45)};
\addplot[mark=diamond] coordinates          {(200,91)(400,95.5)(600,95.8)(800,96.25)(1000,97)(1200,95.9)};
\addplot[mark=pentagon] coordinates         {(200,89.5)(400,93.5)(600,93)(800,92.6)(1000,94)(1200,95)};
\addplot[mark=Mercedes star] coordinates    {(200,95.5)(400,96.5)(600,98.3)(800,98.5)};
\addplot[mark=otimes] coordinates           {(200,89.5)(400,94.2)(600,93.5)(800,93.5)};

\legend{EUR EN$\rightarrow$DE,EUR DE$\rightarrow$EN,EUR FR$\rightarrow$EN,EUR EN$\rightarrow$FR,HAN FR$\rightarrow$EN,
HAN EN$\rightarrow$FR,LIT EN$\rightarrow$DE,LIT DE$\rightarrow$EN,LIT FR$\rightarrow$EN,LIT EN$\rightarrow$FR}
\end{axis}
\end{tikzpicture}
\caption{Supervised classification accuracy as function of number of chunks using function words.}
\label{fig:classification-num-chunks}
\end{figure*}
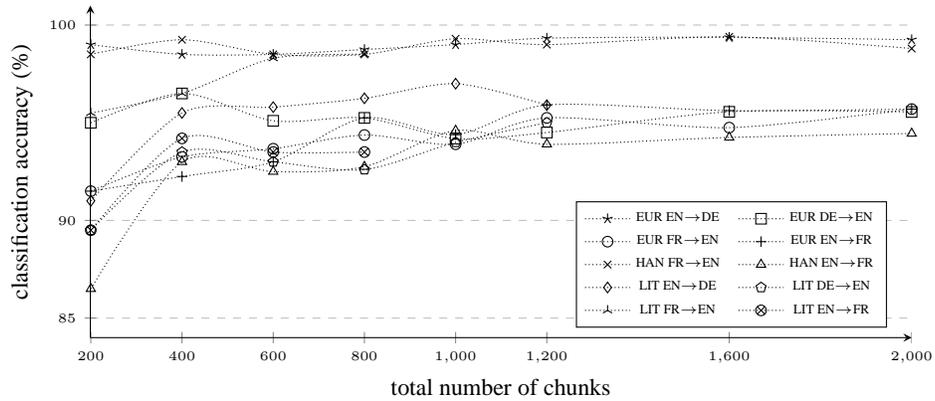

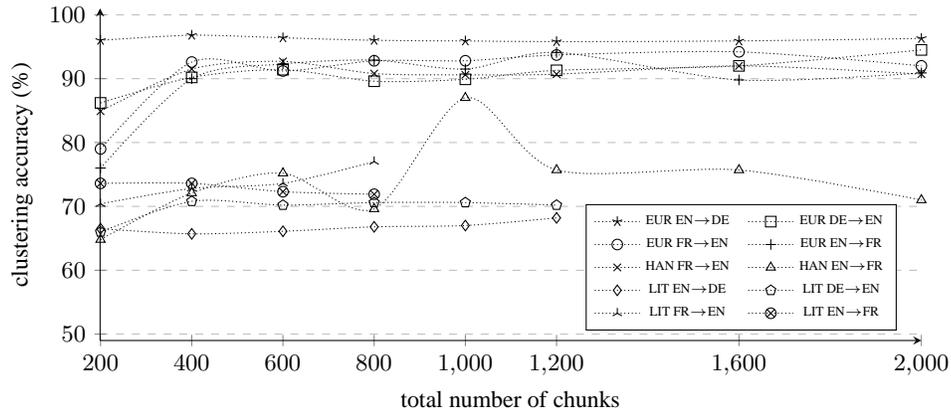
\begin{figure*}[hbt]
\centering
\pgfplotsset{every axis plot/.append style={smooth,black!100,densely dotted}}
\begin{tikzpicture}
\begin{axis}[
    axis lines=middle,
    height=6cm, width=12.5cm,
    xlabel={total number of chunks},
    ylabel={clustering accuracy (\%)},
    ylabel near ticks,
    xlabel near ticks,
    xmin=199, xmax=2000,
    ymin=49,  ymax=101,
    xtick={200,400,600,800,1000,1200,1600,2000},
    ytick={50,60,70,80,90,100},
    ymajorgrids=true,
    grid style=dashed,
    legend pos=south east,
    legend style={/tikz/column 2/.style={column sep=5pt,},font=\tiny},
    legend columns=2,
    mark options={solid,black},
    ]
\addplot[mark=star] coordinates             {(200,96)(400,96.8)(600,96.4)(800,96)(1000,95.9)(1200,95.8)(1600,95.9)(2000,96.3)};
\addplot[mark=square] coordinates           {(200,86.2)(400,90.2)(600,91.4)(800,89.6)(1000,89.9)(1200,91.3)(1600,92)(2000,94.5)};
\addplot[mark=o] coordinates                {(200,79)(400,92.6)(600,91.2)(800,92.8)(1000,92.8)(1200,93.7)(1600,94.2)(2000,92)};
\addplot[mark=+] coordinates                {(200,76)(400,90)(600,92.3)(800,92.9)(1000,91.5)(1200,94.1)(1600,89.8)(2000,90.9)};
\addplot[mark=x] coordinates                {(200,84.9)(400,91.5)(600,92.7)(800,90.8)(1000,90.6)(1200,90.7)(1600,92)(2000,90.7)};
\addplot[mark=triangle] coordinates         {(200,64.8)(400,72.1)(600,75.2)(800,69.6)(1000,87.0)(1200,75.7)(1600,75.7)(2000,71)};
\addplot[mark=diamond] coordinates          {(200,66.5)(400,65.7)(600,66.1)(800,66.8)(1000,67)(1200,68.2)};
\addplot[mark=pentagon] coordinates         {(200,66)(400,70.8)(600,70.2)(800,70.6)(1000,70.6)(1200,70.2)};
\addplot[mark=Mercedes star] coordinates    {(200,70.4)(400,72.8)(600,73.6)(800,77)};
\addplot[mark=otimes] coordinates           {(200,73.6)(400,73.6)(600,72.3)(800,71.9)};

\legend{EUR EN$\rightarrow$DE,EUR DE$\rightarrow$EN,EUR FR$\rightarrow$EN,EUR EN$\rightarrow$FR,HAN FR$\rightarrow$EN,
HAN EN$\rightarrow$FR,LIT EN$\rightarrow$DE,LIT DE$\rightarrow$EN,LIT FR$\rightarrow$EN,LIT EN$\rightarrow$FR}
\end{axis}
\end{tikzpicture}
\caption{Clustering accuracy as function of number of chunks using function words.}
\label{fig:clustering-num-chunks}
\end{figure*}

Next we tested (supervised and unsupervised) classifiers' sensitivity by varying the number of chunks that are subject to classification. We used FW (one of the best performing, content-independent features) in these experiments. We excluded TED and politics datasets from these experiments due to their small size; the results for the literature corpus are limited by the amount of available data in this dataset. Figures~\ref{fig:classification-num-chunks} and~\ref{fig:clustering-num-chunks} report supervised and unsupervised classification accuracy as function of number of chunks used for this task.

Supervised classification accuracy remains stable when the number of chunks used for classification decreases. Evidently, as few as 200 (100 on each side) chunks are sufficient for excellent classification in most cases. Clustering results demonstrate similar pattern: the vast majority of datasets preserve perfectly stable performance when the number of chunks decreases. A single exception is Hansard French (O + T from English), that exhibits results with considerable variance; we attribute these fluctuations to the random choice of samples, subject for clustering.

Unsupervised classification is inherently sensitive procedure, thus the stable accuracy obtained by the majority of sub-corpora implies high reliability and applicability of the clustering procedure to scenarios where only little data are available.

\section{Conclusion}
We present diverse parallel bilingual English-French and English-German corpora with accurate indication of the translation direction. To evaluate the quality of the corpus, we carried out series of experiments across all sub-corpora, using both supervised and unsupervised methodologies and various feature types. This is the first work (to the best of our knowledge) employing unsupervised classification across multiple languages and diverse registers, and the encouraging results stress the applicability of this methodology, leveraging further research in this field.

It has been shown in a series of works \cite{Kurokawa:etal:2009,lembersky-ordan-wintner:2012:EACL2012,lembersky-ordan-wintner:CL2014,lembersky-ordan-wintner:2011:EMNLP,lembersky-ordan-wintner:CL2012,naama:2015} that awareness to translationese has a positive effect on the quality of SMT. Parallel resources presented in this work enable exploitation of \emph{bilingual} information for the task of identification of translationese. More precisely, the datasets that we compiled can be used for the task of identifying the translation direction of parallel texts; task that enjoys growing interest in recent years \cite{eetemadi-toutanova:2014,eetemadi:toutanova:2015}.

The potential value of this work leaves much room for further exploratory and practical activities. Our future plans include extending this set of corpora to additional domains and languages, as well as exploitation of bilingual information for highly accurate identification of translationese at small text units, eventually, at the sentence level.

\section*{Acknowledgments}
This research was supported by a grant from the Israeli Ministry of Science and Technology. We are grateful to Noam Ordan for much advice and encouragement. We also thank Sergiu Nisioi for helpful suggestions. We are grateful to Philipp Koehn for making the Europarl corpus available; to Cyril Goutte, George Foster and Pierre Isabelle for providing us with an annotated version of the Hansard corpus; to Fran\c{c}ois Yvon and Andr\'{a}s Farkas\footnote{\url{http://farkastranslations.com}} for contributing their literary corpora; and to the TED OTP team for sharing TED talks and their translations. We thank also Raphael Salkie for sharing his diverse English-German corpus.

\newpage
\bibliographystyle{splncs}
\bibliography{all}

\end{document}